\def\year{2022}\relax
\documentclass[letterpaper]{article} 
\usepackage{aaai22}  
\usepackage{times}  
\usepackage{helvet}  
\usepackage{courier}  
\usepackage[hyphens]{url}  
\usepackage{graphicx} 
\usepackage{natbib}  
\usepackage{caption} 
\DeclareCaptionStyle{ruled}{labelfont=normalfont,labelsep=colon,strut=off} 
\usepackage{algorithm}
\usepackage{algorithmic}
\usepackage{amsmath}
\usepackage{xcolor}
\usepackage{graphicx}
\usepackage{amsmath}
\usepackage{csquotes}
\usepackage{newtxmath}
\usepackage[version=4]{mhchem}
\usepackage{siunitx}
\usepackage{longtable,tabularx}
\usepackage{bigints}
\usepackage{bm}
\usepackage{float}
\usepackage{listings}
\floatstyle{ruled}
\newfloat{listing}{tb}{lst}{}
\floatname{listing}{Listing}
\title{Symmetry Detection in Trajectory Data for More Meaningful Reinforcement Learning Representations}
\author {
    Marissa D'Alonzo,
    Rebecca Russell
}
\affiliations {
    The Charles Stark Draper Laboratory, Inc. \\
    mdalonzo@draper.com, rrussell@draper.com
}

\begin{document}

\maketitle


\begin{abstract}
Knowledge of the symmetries of reinforcement learning (RL) systems can be used to create compressed and semantically meaningful representations of a low-level state space. We present a method of automatically detecting RL symmetries directly from raw trajectory data without requiring active control of the system. Our method generates candidate symmetries and trains a recurrent neural network (RNN) to discriminate between the original trajectories and the transformed trajectories for each candidate symmetry. The RNN discriminator's accuracy for each candidate reveals how symmetric the system is under that transformation. This information can be used to create high-level representations that are invariant to all symmetries on a dataset level and to communicate properties of the RL behavior to users. We show in experiments on two simulated RL use cases (a pusher robot and a UAV flying in wind) that our method can determine the symmetries underlying both the environment physics and the trained RL policy.
\end{abstract}

\section{Introduction} \label{sec:intro}

With the advancement of machine learning, the capabilities of autonomous agents trained using reinforcement learning (RL) are improving rapidly, often outperforming humans and expert autonomous systems. However, a significant stumbling block in the safe deployment of RL comes from the black-box nature of the technology. RL agents are typically uninterpretable and can exhibit behavior unaligned with human expectations. In addition, RL agents often operate in high-dimensional state representations that contain low-level information, making it difficult even for experts to extract relevant information and behavior patterns, a problem known as the readability-performance trade-off \cite{survey2}. Methods that automate the clear communication of interesting and relevant information from machine learning systems to users, also known as ``explainable AI,'' remains an active area of research.

State Representation Learning (SRL) is one approach to explainable AI that aims to automatically craft a lower-dimensional and semantically meaningful representation of a state space by processing high-dimensional raw observation data. SRL representations can encompass both the agents' actions and the resulting environment, and thus are particularly relevant in robotics applications ~\cite{survey1}. Good state representations reflect the concept of a ``simplicity prior,'' which states that for a given task, only a small number of world properties are relevant ~\cite{robotic_prior}. One way to fulfill this requirement is to define a state representation in which RL agent behavior is invariant to its underlying symmetries, thus capturing only meaningful information and consistently representing behavior. In real systems, where data collection can be expensive and difficult, identifying the approximate behavioral symmetries of the RL system is a challenging problem.

\begin{figure}[t]
    \centering
    \includegraphics[width=0.35\textwidth]{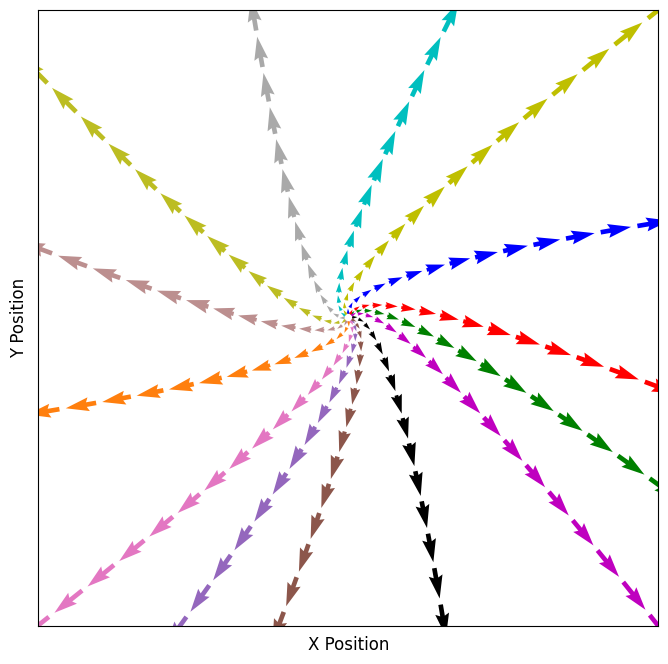}
    \caption{An approximate rotational symmetry in two dimensions. While a human can easily see that all the trajectories exhibit similar behavior, this is difficult to automatically detect.}
    \label{fig:symmetry}
\end{figure}

Figure~\ref{fig:symmetry} illustrates an example of a trajectory symmetry that can be detected at the dataset level. In the raw representation, the trajectories are not close to each other, but a human can see that all the trajectories are similar up to a rotation about the origin. This symmetry is difficult to automatically determine, particularly if it is inexact. Once the symmetry is detected, the trajectories can be mapped to the same representation. As the number of dimensions in the raw state representation increases beyond two, even a human will struggle to identify trajectory symmetries.

 The goal of this work is to leverage machine learning to automatically identify symmetries in the raw state representation of an autonomous agent without actively collecting direct pair-to-pair trajectory data. Once these symmetries are identified, we can craft a state representation that fulfills the simplicity prior by being invariant to symmetries in both the agent behavior and the physical world. This representation communicates to the user what information about the agent is most meaningful and can speed up SRL by removing unnecessary information from the input data. In addition, the RL policy may introduce or break symmetries that are present in the physical system dynamics. The comparison of symmetries and asymmetries present in the system between a random policy and the RL policy is useful to the end user by providing additional context to potentially counter-intuitive behaviors and is also helpful to the developer for debugging purposes.

\begin{figure*}[th!]
\centering
    \includegraphics[width=0.9\textwidth]{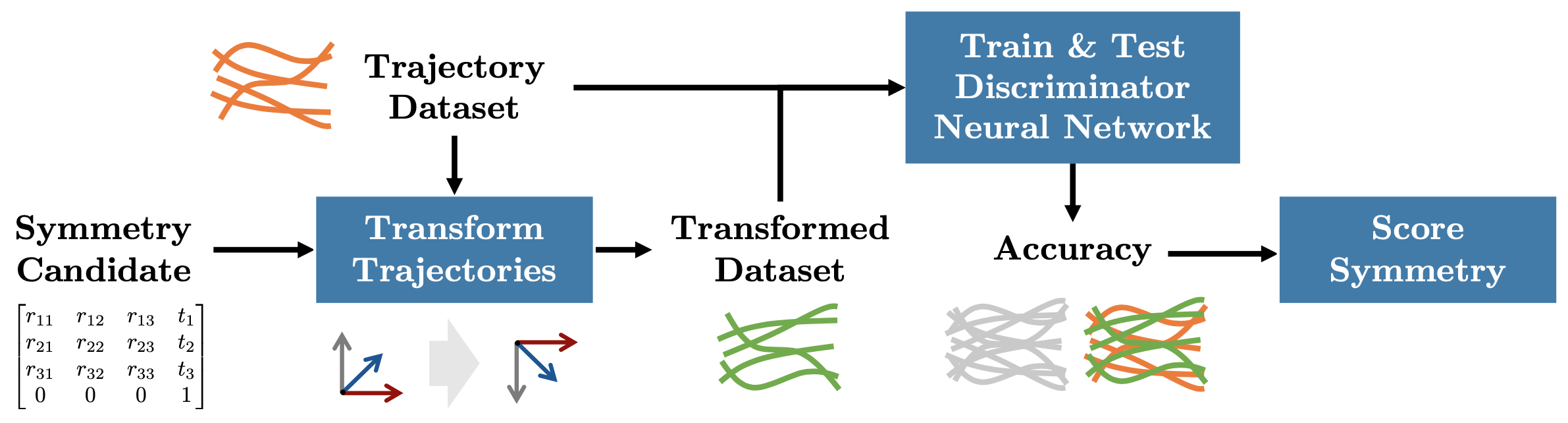}
    \caption{An overview of our symmetry detection pipeline. We train a discriminator neural network to distinguish between the original trajectories and trajectories transformed under a candidate symmetry. The accuracy of the discriminator on test data allows us to score the symmetry of the trajectories under the given transformation.}
    \label{fig:symmetryOverview}
\end{figure*}

We investigate the automation of symmetry detection by training a neural network to discriminate between the original trajectories and transformed trajectories under a candidate symmetry. The neural network being unable to distinguish between the two data distributions suggests the trajectories are symmetric under the respective transformation. We demonstrate our approach on two simulated RL problems: a pusher robot and a UAV flying in wind.

\section{Background and Related Work}

Automating the interpretation of black-box systems or ``explainable AI'' remains an open area of research. Even the definition of ``interpretability'' is up for debate \cite{Interpretable}, with some defining it as the ability of experts to interpret the results of a black-box system, while others consider it to be the ability of a layperson to understand why the system has made a particular decision. In this work, we consider ``interpretability'' to be the communication of semantically meaningful and relevant information about the behavior of a autonomous agent to a human operator. This information should be digestible by the user of the given system, most likely a person who has received minimal training on the system and would not be considered an expert. The user should be able to leverage the provided information to make decisions about the state of the mission in-situ and perform post-hoc analysis of the agent behavior.

There are various sub-areas of interest within the field of explainable AI, including algorithms that simultaneously learn a policy and its explanation through reward decomposition ~\cite{Juozapaitis2019ExplainableRL} and action influence models ~\cite{causal}, but these provide explanations of specific actions, not the overall agent behavior for a given scenario. Other efforts have focused on post-hoc explainability by using methods such as saliency maps ~\cite{saliency} and symmetry-based disentagled representation learning ~\cite{disentangled_symmetry} to learn representations on pre-trained agents, but these methods focus on extracting semantically meaningful information from images, not trajectories. Some research methods focus on the creation of abstract trajectories relative to the task specification \cite{abstractTraj}, which provide semantically interesting and relevant results, but these methods still require significant hand-engineered effort. Temporal Logic ~\cite{LTL, LTL2}, in which a predefined set of modal operators is used to describe sequences based off temporal properties, has also been used to interpret black-box systems, but this approach suffers from poor scalability.

State Representation Learning (SRL) is a sub-area of explainable AI that works to craft a lower-dimensional and (in most cases) semantically meaningful representation of a state space by processing high-dimensional raw observation data, which can include both agent actions and the resulting environment. Many researchers create a lower-dimensional representation by training an autoencoder that performs denoising or reconstruction of the original state vector, but the results of these methods are not semantically interesting ~\cite{DBLP:journals/corr/FinnTDDLA15, autoencoder} and thus are not suitable for our purposes. Other SRL methods leverage information the structure imposed on the system by physics, also called the ``robotic prior.'' Research in this realm ~\cite{robotic_prior, robotic_prior2} defines six priors---simplicity, temporal coherence, proportionality, causality, and repeatability. The simplicity prior states that for a given task, only a small number of world properties are relevant. Previous research ~\cite{robotic_prior} assumes that the user will know which properties are relevant for a given task ahead of time and focuses on calculating the other priors, but in our case, this is not true. Even if the symmetries of the underlying physical system are well understood, the RL policy could add or remove symmetries and it is prohibitively expensive to gather the amount of data necessary to confirm these symmetries in a real system. Our goal is to automate the learning of the relevant world properties by discovering symmetries inherent in the data by extending research conducted in the field of high-energy physics in works such as ~\cite{SymmetryGAN} to work with RL generated trajectory data.

\section{Methods}

\subsection{Experimental Approach}

We begin with a set of transformations for candidate symmetries, the selection of which is discussed below, to evaluate for the given RL tasking. To determine if the trajectories are symmetrical under a given transformation, that transformation is applied to the original trajectory data to create a new dataset. The original dataset and the transformed dataset are then labeled accordingly and used to train the ``discriminator'' neural network. This network attempts to learn the difference between the two labeled sets. If, after training, the discriminator is unable to distinguish between the two datasets, the candidate is considered a symmetry. The datasets are considered indistinguishable if the classification accuracy of the test set is near 50 percent. A discriminator neural network is trained for each transformation examined. Figure \ref{fig:symmetryOverview} provides an overview diagram of our system.

\subsection{Network Architecture} \label{sec:arch}
The neural network was implemented in PyTorch \cite{pytorch} and consisted of three layers. A linear layer encoded the state dimensions into 8 dimensions. A gated recurrent unit (GRU) served as the recurrent section of the neural network. This architecture was selected due to the time-varying nature of the data, as well as the network's ability to use its memory to learn a compressed representation of the full history while also being parameter and data efficient. The GRU contained a single layer of 32 hidden units. A second linear layer compressed the maxpooled outputs of the GRU down to a single logit output to be used for binary classification. While these hyperparameters were appropriate for the RL datasets used in our experiments, they can be tuned to maximize validation set accuracy on other datasets as needed.

\subsection{Candidate Transformations} \label{sec:transformations}

For this problem, we considered potential symmetries in 2D and 3D Euclidean space. While an infinite number of transformations could be examined, we constrained ourselves to a basic set of simple transformations that could be used to compose other, more complex transformations: $C_2$ to $C_8$ rotations, general (any angle) rotations, reflections, and general (any distance) translations. These transformations are tested with respect to any meaningful geometric axes in the system. Future work discussed in the conclusion will discuss the potential automation of transformation selection.

It is also important that the initial state distribution of the dataset is symmetric under these transformations in order for our approach to be a good test of the symmetry of the RL policy. If the two initial state distributions differ, it is trivial for the network to discriminate. To satisfy this requirement, we sampled the initial positions uniformly from a region within a given radius of the origin. If the initial state distribution in the data does not satisfy this requirement, the data should be downsampled so that it does.

\begin{table}[h!]
    \caption{Overview of the transformations examined}
    \begin{tabular}{ |c|c| } 
         \hline
        \textbf{Transformation} & \textbf{Description}  \\
        \hline
        Translation & Randomly add or  \\
          & subtract value within domain bounds \\
        \hline
        General Rotation & Randomly rotate  \\
         & the value between -360 $^{\circ}$ and 360 $^{\circ}$ \\
        \hline
        Reflection & Reflect the value around the given axis \\
        \hline
        $C_{2}$ Rotation & Rotate the values 180 $^{\circ}$ \\ 
        \hline
        $C_{3}$ Rotation & Rotate the values 120 $^{\circ}$ \\ 
        \hline  
        $C_{4}$ Rotation & Rotate the values 90 $^{\circ}$ \\
        
        \hline
        $C_{6}$ Rotation & Rotate the values 60 $^{\circ}$ \\
        \hline
        $C_{8}$ Rotation & Rotate the values 45 $^{\circ}$ \\
        \hline
    \end{tabular}
    \label{table:transformations}
\end{table}

\section{Experiments}
The following sections detail the RL use cases examined, the experimental setup, and results.

\subsection{Pusher Robot RL Use Case}

\begin{figure}[h!]
    \centering
    \includegraphics[width=0.4\textwidth]{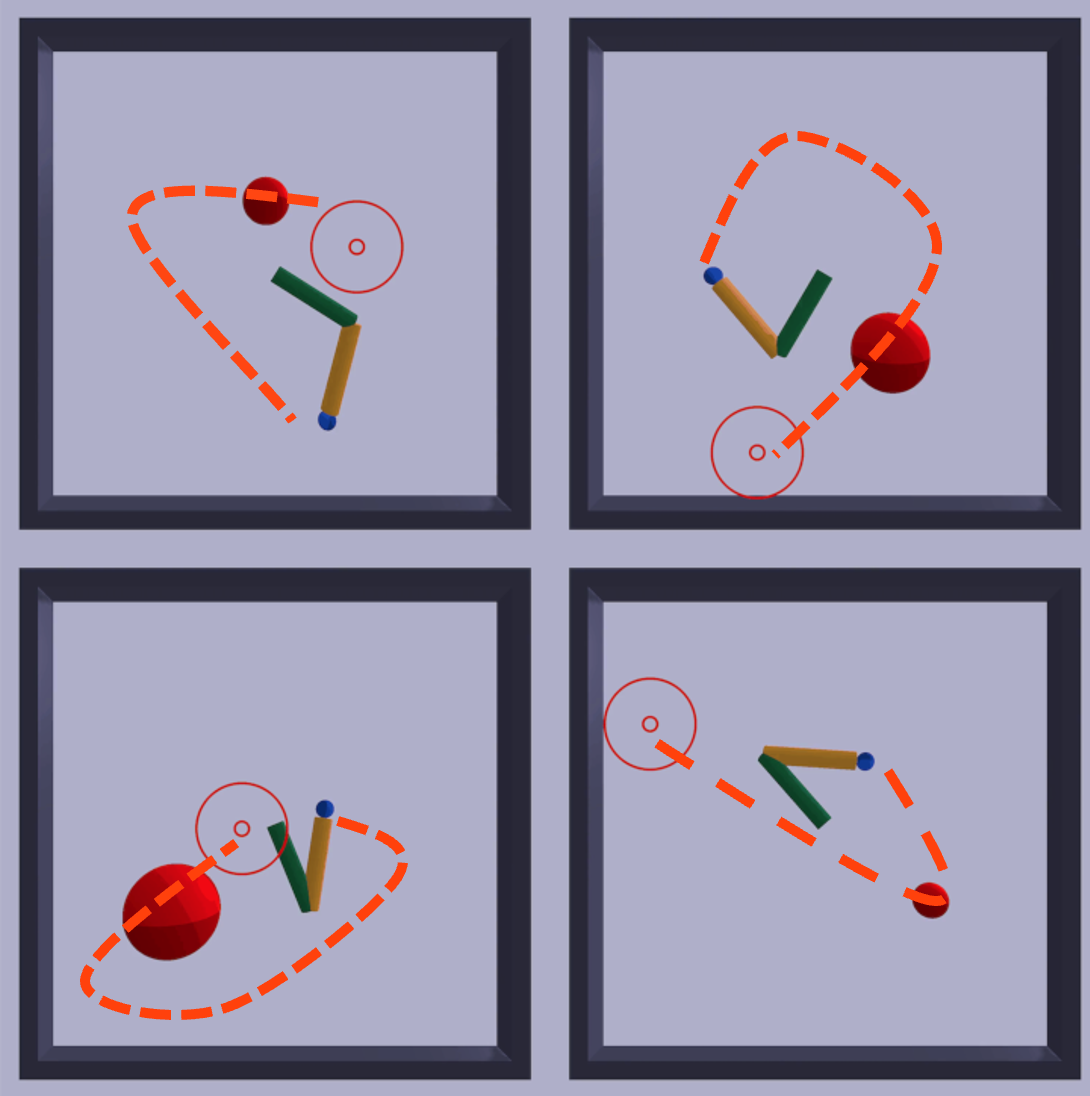}
    \caption{Four randomly-sampled pusher robot trajectories. The arm attempts to push the ball to the target location indicated by the red circles. The dotted orange line shows the trajectory of the robot hand.}
    \label{fig:pusher}
\end{figure}

The pusher robot is a PyBullet RL problem where the goal is to use a robotic arm to push a ball of varying size to a target location. It is a modified version of the reacher environment from ~\cite{pusher}. The relevant state space contains 14 variables defining the positions, orientations, velocities, and rotations of the ball and arm components, as well as the position of the target. Figure \ref{fig:pusher} shows several different initial conditions in which the arm attempts to push the ball to the target location indicated by the red circles while dealing with randomized target position, ball radius, and robotic pose. 10k trajectories are generated with a random policy which aims to simply move the robotic arm to examine symmetries the physical environment. Two additional policies trained to complete the goal are also examined to discover symmetries in the agent behavior, with 32k and 10k samples being gathered, respectively. Both are trained according using the process outlined in ~\cite{sciTech}.

\subsection{UAV in Wind RL Use Case}
Our second RL use case is a Gazebo ~\cite{gazebo} simulation of a UAV (Unmanned Aerial Vehicle) flying in variable wind conditions to given target locations in 3D dimensions. The mission area is a $500\times500$ meter zone with two randomly placed targets centered around the initial $x$-$y$ position of the UAV. The UAV itself is placed at a randomly selected altitude and is given 60 seconds to reach at least one target before the end of its battery life. The wind direction and magnitude, temperature, and payload weight are all randomly varied. The RL state representation is 24 variables defining the pose and change in pose of the UAV, temperature, wind direction, and payload mass. First, data is collected with a random policy in order to examine the symmetries in the physical environment. Next, data is gathered using a policy trained to accomplish the above mission as outlined in ~\cite{sciTech}. 164k randomly generated trajectories and 10k planned trajectories are examined.

\begin{figure}[t]
    \centering
    \includegraphics[width=0.48\textwidth]{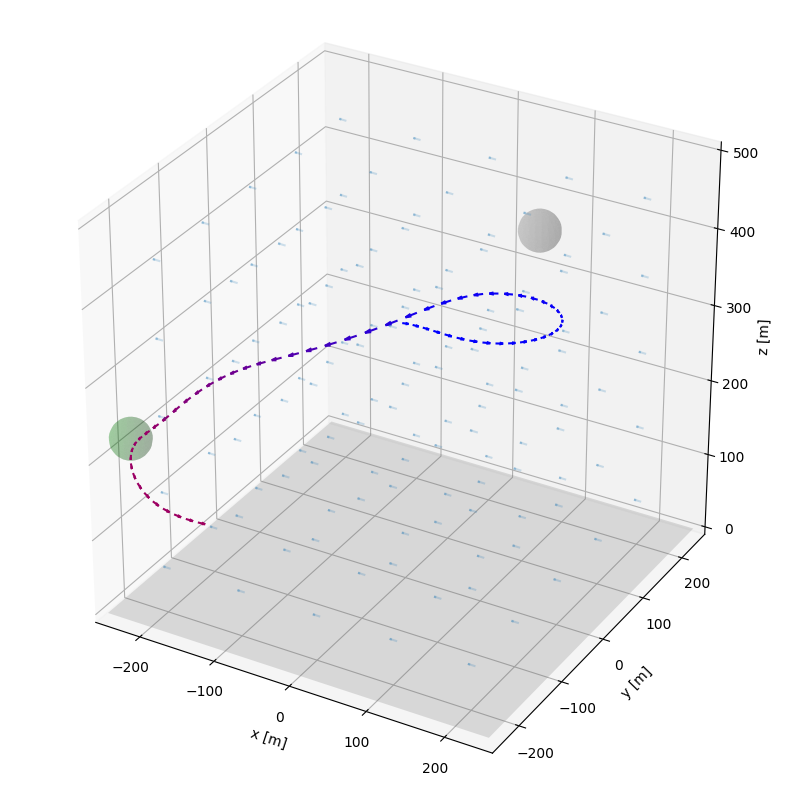}
    \caption{Example visualization of the UAV use case. The arrows indicate the path of the UAV, with the coloring indicating battery level. Blue indicates a full battery, while red indicates no battery. The repeating blue arrows represent wind direction. The two spheres indicate the position of the targets. A green sphere indicates the target was reached by the UAV, while a grey sphere was not.}
    \label{fig:uav}
\end{figure}

\subsection{Experimental Setup}
A separate experiment is performed for all transformations, datasets, and RL policies examined. The discriminator recurrent neural network is trained using the Adam optimizer and binary cross entropy loss.

\subsection{Results}
\subsubsection{Pusher Robot Results}                                          
The transformations outlined above are examined for three reinforcement learning policies---two trained to complete the mission of pushing the ball to the target and one random policy which simply moves the robotic arm according to actions selected from a Gaussian distribution. The examination of data with random actions aims to determine symmetries of the physical system, while the planned data aims to find the behavioral symmetries of the RL policies. 

The relevant transformations are applied to the pusher robot arm angles, arm velocities, ball position and ball velocity. In the random case, no target is considered, but for the RL policy datasets, the transformation is applied to the target position. We consider the coordinate frame in which the robot base is centered at the origin and the $x$ and $y$ axes are perpendicular to the square walls of the pen surrounding the robot area.

\begin{table*}[t]
\centering
    \caption{The accuracy reported by the discriminator for each transformation, dataset and planner for the pusher robot. Values in \textbf{bold} indicate a symmetry because the network is unable to distinguish between the original and transformed data sets.}

    \begin{tabular}{ |c|c|c|c| } 
    \hline
        \textbf{Transformation} & \textbf{Random Policy Accuracy ($\%$)}  & \textbf{RL Policy \#1 Accuracy}  ($\%$)   & \textbf{RL Policy \#2 Accuracy}  ($\%$) \\
        \hline
        $C_{2}$ Rotation & \textbf{50} & \textbf{50} & \textbf{50} \\ 
        \hline
        $C_{3}$ Rotation & 60 & 78 & 64 \\ 
        \hline
        $C_{4}$ Rotation & \textbf{50} & \textbf{50} & \textbf{50} \\   
        \hline
        $C_{6}$ Rotation & 59 & 77 & 67 \\ 
        \hline
        $C_{8}$ Rotation & 59 & 79 & 68 \\ 
        \hline
        General Rotation & 51 & 64 & 58 \\ 
        \hline
        X Reflection & \textbf{50} & 55 & \textbf{50} \\ 
        \hline
        Y Reflection & \textbf{50} & 56 & \textbf{50} \\ 
        \hline
        X Translation & 90 & 81 & 67 \\ 
        \hline
        Y Translation & 82 & 96 & 72 \\ 
        \hline
    \end{tabular}
    \label{table:Results}
\end{table*}

When analyzing the randomly generated data, the discriminator is easily able to distinguish between the original and the translated trajectories. This indicates that no symmetry is found. Symmetries are discovered in reflection, $C_{2}$, and $C_{4}$ rotations, as evidenced by the low discriminator accuracy, but not $C_{3}$, $C_{6}$, $C_{8}$, or general rotations. This is expected, as the non-symmetrical transformations can result in ball positions outside the values encompassed by the pen and different physical affects when bouncing off of the walls.

When examining the data generated using the two trained RL policies, similar behaviors are discovered: symmetry is preserved in $C_{2}$ and $C_{4}$ rotations, but not translation, general rotation, or $C_{3}$, $C_{6}$ and $C_{8}$ rotations. In RL policy \#1, $x$ and $y$ reflection are no longer considered symmetries, as shown by the increased discriminator accuracy. However, this is not true of RL policy \#2, in which reflection symmetry is preserved. This appears to be due to a surprising bias we have observed in Policy \#1 that causes it to prefer moving the robot arm counterclockwise over clockwise. This shows that the addition of the RL behavior can break certain environment symmetries, but these effects can be dependent on the planner parameters and training. In addition, it is interesting to note that the discriminator model has higher classification accuracy for the RL data when compared to the random trajectory data, which suggests that the RL policy enhances the physical non-symmetries found in the system. This is logical for this use case because the RL policy often leverages the physics of the system by bouncing the ball against the walls. These trajectories with more distinct physical interactions contain more information about the symmetry of the system.

\begin{figure*}[h!]
\centering
    \includegraphics[width=0.9\textwidth]{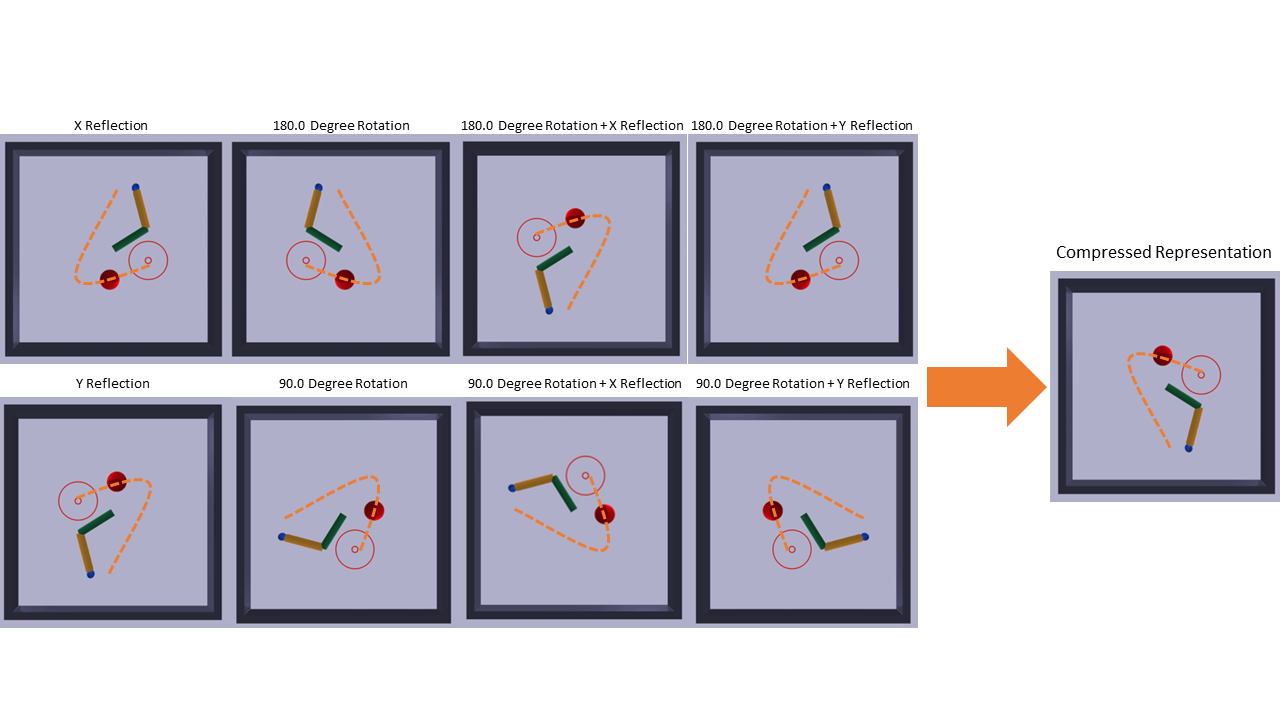}
    \caption{An example of the benefits of automatic symmetry detection---the left shows eight seemingly distinct pusher start positions. The title of each image shows the transformation that must be applied to get the position on the right. Due to the identified $C_2$ rotation, $C_4$ rotation, and reflection symmetries, these states can all can be considered identical.}
    \label{fig:pusher_mapping}
\end{figure*}

The identification of these symmetries can allow us to map multiple, seemingly distinct, pusher trajectories into a single representation, as shown in Figure \ref{fig:pusher_mapping}. Representations where these trajectories are equivalent better communicate to the user what the RL policy considers important and can be used to reduce the dimensionality of the state representation.

\subsubsection{UAV in Wind Results}

The transformations outlined above are examined for both data generated under a random policy and data generated using a model-based RL policy that aims to have the UAV complete the task outlined above.

The relevant transformations are applied to the UAV position and velocity and wind direction. In the random policy case, no target is considered, and in the RL policy case, the transformation is applied to the target positions. The UAV orientation and angular velocity were not considered because the long-horizon trajectory dynamics are adequately captured from the position and velocity alone. Battery, temperature, and payload were fed to the discriminator, but no transformations were applied since they are non-geometric quantities. We consider the vehicle frame, in which the $x$ axis points north, the $y$ axis points east, and the $z$ axis points towards the Earth's center.

When examining the data generated with the random policy, the discriminator can easily distinguish between the original data set and the data set reflected along the $z$ axis, while it cannot distinguish between those reflected along the $x$ or $y$ axes. This indicates the existence of symmetries in the $x$ and $y$ axes but not the $z$ axis. Similarly, the discriminator can clearly classify all trajectories rotated around the $x$ and $y$ axes, but not $z$ axis. These results are logical because the $z$ axis is affected by gravity while the others are not. Results for the $C_{*}$ rotations were equivalent to the general axes rotations and thus were excluded from the table.

When examining the data generated with the trained RL policy, the same symmetries are observed in all transformations. In addition, it is interesting to note that unlike the pusher robot, the discriminator model has similar classification accuracy for both the RL policy and the random trajectory data, which indicates that, in this scenario, the RL policy does not enhance the physical non-symmetries in the system.
\begin{table*}[t]
\centering
    \caption{The accuracy reported by the discriminator for each transformation, data set and policy for the UAV use case. Values in \textbf{bold} indicate a symmetry because the network is unable to distinguish between the original and transformed data sets}

    \begin{tabular}{ |c|c|c| } 
    \hline
        \textbf{Transformation} & \textbf{Random Policy Accuracy ($\%$)}  & \textbf{RL Policy Accuracy}  ($\%$) \\
        \hline
        General X Rotation & 88 & 93 \\
        \hline
        General Y Rotation & 89 & 92 \\
        \hline
        General Z Rotation & \textbf{48} & \textbf{49}\\
        \hline
        X Reflection & \textbf{50} & \textbf{50} \\
        \hline
        Y Reflection & \textbf{50}& \textbf{50} \\
        \hline
        Z Reflection & 99 & 100 \\ 
        \hline
        X Translation &\textbf{50} &\textbf{50}\\
        \hline
        Y Translation & \textbf{50} & \textbf{50}\\
        \hline
        Z Translation & 53 & 94 \\ 
        \hline

    \end{tabular}
    \label{table:Results}
\end{table*}

\section{Conclusion} \label{sec:conclusion}

In this paper, we present a method for automating the discovery of symmetries in both the environment and the system behavior of an autonomous RL agent. The determination of these symmetries can be used to identify functionally similar trajectories. We apply a candidate transformation to a dataset, then train a recurrent neural network to distinguish between the original and transformed dataset. If the network is not able to distinguish between them, we say the system is symmetric under that transformation. We present results for a pusher robot RL and a UAV in wind use case. In both cases, we are able to identify symmetries created by both the environment and the RL policy and use these to identify functionally equivalent trajectories.

Future work on this topic should focus on two key areas: the automation of the selection of the transformation state space and the automatic downselection of the state representation. This work uses expert knowledge to define the transformations examined and create the invariant state representation, but future work should aim to automate this process. These could both be accomplished by imitating the purpose of a generative adversarial network (GAN) and attempting to learn a generator of the parameters of transformations that fool the discriminator network. By enabling us to create more meaningful representations of RL trajectories, these approaches may allow us to communicate the patterns of RL behavior to human users and make RL more safe and practical.

\appendix

\bibliography{aaai22.bib}

\section{Acknowledgments}
This material is based upon work supported by the Defense Advanced Research Projects Agency (DARPA) under Contract No. HR001120C0032. Any opinions, findings and conclusions or recommendations expressed in this material are those of the author(s) and do not necessarily reflect the views of DARPA. 
\end{document}